%% file: main.tex
\pdfoutput=1

\documentclass[11pt]{article}

\usepackage[]{EMNLP2023}

\usepackage{times}
\usepackage{latexsym}

\usepackage[T1]{fontenc}

\usepackage[utf8]{inputenc}

\usepackage{microtype}

\usepackage{inconsolata}

\usepackage{amssymb}
\usepackage{amsmath}
\usepackage{amsfonts}
\usepackage{graphicx}
\usepackage{threeparttable}
\usepackage{booktabs}

\begin{document}

\title{Quantifying Character Similarity with Vision Transformers}
\author{Xinmei Yang$^{1}$, Abhishek Arora$^{2}$, Shao-Yu Jheng$^{2}$, Melissa Dell$^{2, 3^\ast}$ \\
\normalsize{$^{1}$Renmin University; Beijing, China.}\\
\normalsize{$^{2}$Harvard University; Cambridge, MA, USA.}\\
\normalsize{$^{3}$National Bureau of Economic Research; Cambridge, MA, USA.}\\
\normalsize{$^\ast$Corresponding author:  melissadell@fas.harvard.edu.}
}
\maketitle

\begin{abstract}
    Record linkage is a bedrock of quantitative social science, as analyses often require linking data from multiple, noisy sources. Off-the-shelf string matching methods are widely used, as they are straightforward and cheap to implement and scale. Not all character substitutions are equally probable, and for some settings there are widely used handcrafted lists denoting which string substitutions are more likely, that improve the accuracy of string matching. However, such lists do not exist for many settings, skewing research with linked datasets towards a few high-resource contexts that are not representative of the diversity of human societies. This study develops an extensible way to measure character substitution costs for OCR'ed documents, by employing large-scale self-supervised training of vision transformers (ViT) with augmented digital fonts. For each language written with the CJK script, we contrastively learn a metric space where different augmentations of the same character are represented nearby. In this space, homoglyphic characters - those with similar appearance such as ``O'' and ``0'' - have similar vector representations. Using the cosine distance between characters' representations as the substitution cost in an edit distance matching algorithm significantly improves record linkage compared to other widely used string matching methods, as OCR errors tend to be homoglyphic in nature. Homoglyphs can plausibly capture character visual similarity across \textit{any} script, including low-resource settings. We illustrate this by creating homoglyph sets for 3,000 year old ancient Chinese characters, which are highly pictorial. Fascinatingly, a ViT is able to capture relationships in how different abstract concepts were conceptualized by ancient societies, that have been noted in the archaeological literature. 
\end{abstract}

\section{Introduction}
\label{sec:intro}

Many quantitative analyses in the social sciences - as well as government and business applications - require linking information from multiple datasets. 
For example, researchers and governments link historical censuses, match hand-written records from vaccination campaigns to administrative data, and de-duplicate voter rolls. 
The sources to be linked often contain noise, particularly when they were created with optical character recognition (OCR). 
String matching methods are widely used to link entities across datasets, as they are straightforward to implement off-the-shelf and can be scaled to massive datasets \cite{binette2022almost, abramitzky2021automated}. 

Most simply, approximate string matching methods count the number of edits (insertions, deletions, and substitutions) to transform one string into another \cite{levenshtein1966binary}. Another common approach computes the similarity between $n$-gram representations of strings, where $n$-grams are all substrings of length $n$ \cite{okazaki2010simple}. 

In practice, not all string substitutions are equally probable, and efforts to construct lists that vary their costs date back over a century. 
For example, in 1918 Russell and Odell patented Soundex, a sound standardization toolkit that accounts for the fact that census enumerators often misspelled names according to their sound. Together with the updated New York State Identification and Intelligence System \cite{silbert1970world}, it remains a bedrock for linking U.S. historical censuses \cite{abramitzky2021automated}.
Similarly, \citet{masala} adjusts Levenshtein distance to impose smaller penalties for common alternative spellings in Hindi, and the FuzzyChinese package \cite{fuzzychinese} uses strokes as the unit for $n$-grams substring representations, where the strokes for a given character are drawn from an external database \cite{chaizi} covering a subset of the CJK script. Characters sharing strokes are more likely to be matched.

Such methods can perform well in the contexts for which they are tailored but are labor-intensive to extend to new settings, due to the use of hand-crafted features. Low extensibility skews research with linked data - necessary to examine intergenerational mobility, the evolution of firm productivity, the persistence of poverty, and many other topics - towards a few higher resource settings that are not representative of the diversity of human societies.

This study aims to preserve the advantages of string matching methods - simple off-the-shelf implementation and high scalability - while developing an extensible, self-supervised method for determining the relative costs of character substitutions in databases created with OCR.  OCR often confuses characters with their homoglyphs, which have a similar visual appearance (\textit{e.g.} ``0'' and ``O''). Incorporating character visual similarity into string matching can thus plausibly improve record linkage. 
Homoglyphs can be constructed by hand for small script sets such as Latin, as in a psychology literature on literacy acquisition \cite{simpson2013letter}, but for a script such as CJK, containing over 38,000 characters, this is infeasible. 

Following a literature on self-supervision through simple data augmentation for image encoders  \cite{grill2020bootstrap, chen2021empirical, chen2021exploring}, this study uses augmented digital fonts to contrastively learn a metric space where different augmentations of a character have similar vector representations.  The resulting space can be used, with a reference font, to measure the visual similarity of different characters. Figure \ref{fig:font} shows representative examples of how the same characters are rendered very differently across fonts. These different representations form positive examples for the contrastively trained \texttt{HOMOGLYPH} model. This purely self-supervised approach can be extended to any character set, but since creating evaluation data for record linkage is costly, the study focuses on languages written with CJK: Simplified and Traditional Chinese, Japanese, and Korean. 

\input{figs/char_font.tex}

We train on augmentations of the same character - rather than paired data across characters - because a self-supervised approach is more extensible. Paired character similarity data are limited. Unicode maintains a set of confusables - constructed with rule-based methods - but for CJK the only confusables are structurally identical characters with different Unicode codepoints. Despite a large post-OCR error correction literature  \cite{10.1162/tacl_a_00379, 10.1145/3453476, artidigh20}, there is also limited ground truth data about the types of errors that OCR makes across architectures, languages, scripts, layouts, and document contexts.  

Using the cosine distance between two characters as the substitution cost within a Levenshtein edit distance framework \cite{levenshtein1966binary} improves record linkage with 1950s firm level data about Japanese supply chains \cite{pr, teikoku}, relative to other string matching methods. The study also compares to end-to-end deep learning methods for record linkage. While these methods can outperform string matching, the data required for them are not always available and technical requirements for implementation are higher, explaining why string matching methods predominate in social science applications. Homoglyphic matching is a cheap and extensible way to improve these predominant methods.

Because creating annotated ground truth data is costly, we provide additional evaluations using synthetically generated data. We augment image renders of place and firm names written with different fonts, for Simplified and Traditional Chinese, Japanese, and Korean character sets. We then OCR two different views of each entity with different OCR engines - EasyOCR and PaddleOCR - that use very different architectures. The different augmentations and OCR engines lead to different text string views of the same entity with high frequency. We then link these using string matching methods. Homoglyphic matching outperforms other widely used string matching techniques for all four scripts. Our \texttt{HomoglyphsCJK} python package provides a simple, off-the-shelf implementation.\footnote{Package available at \url{https://pypi.org/project/HomoglyphsCJK/}.}

Homoglyphs can be extended to any script. To explore this, we contrastively train a \texttt{HOMOGLYPH} model for ancient Chinese characters, using a database that provides views of the same character from different archaeological sites and time periods \cite{Sinica_Academia}. Ancient characters are much more pictorial than their more abstract, modern equivalents. Fascinatingly, homoglyphs constructed with a ViT for the Shang Dynasty (1600 BC-1045 BC) capture ways in which ancient Chinese society related abstract concepts that have been noted in the archaeological literature (\textit{e.g.} \citet{wang2003}). 

The rest of this study is organized as follows: 
Section \ref{methods} develops methods for learning character similarity and incorporating it into record linkage, and Section \ref{eval} describes the evaluation datasets. 
Section \ref{homomatch} compares the performance of homoglyphic edit distance to other string matching and neural methods for record linkage. Section \ref{ancient} examines extensibility by constructing homoglyphs for ancient Chinese, Section \ref{limits} discusses the limitations of homoglyphs, and Section \ref{conclude} concludes.

\section{Methods} \label{methods}

\subsection{The \texttt{HOMOGLYPH} model}

The \texttt{HOMOGLYPH} model contrastively learns a mapping between character crops and dense vector representations, such that crops of augmentations of the same character are nearby. \texttt{HOMOGLYPH} is trained purely on digital fonts. Figure \ref{fig:font} shows variations of the same characters rendered with different fonts, which form positive examples for training. Variations across fonts are non-trivial, forcing the model to learn character similarities at varying levels of abstraction.

\input{figs/char_near.tex}

We use a DINO (Self-\textbf{Di}stillation, \textbf{No} Labels) pre-trained ViT as the encoder \cite{caron2021emerging}. 
DINO ViT embeddings perform well as a nearest neighbor classifier, making them well-suited for homoglyphic matching.
The model is trained using a Supervised Contrastive loss function  \cite{khosla2020supervised}, a generalization of the InfoNCE loss \cite{oord2018representation} that allows for multiple positive and negative pairs for a given anchor:

\begin{small}
\begin{equation}
\sum_{i \in I} \frac{-1}{|P(i)|} \sum_{p \in P(i)} \log \frac{\exp \left(\boldsymbol{z}_i \cdot \boldsymbol{z}_p / \tau\right)}{\sum_{a \in A(i)} \exp \left(\boldsymbol{z}_i \cdot \boldsymbol{z}_a / \tau\right)}
\end{equation}
\end{small}

where $\tau$ is a temperature parameter (equal to 0.1), $i$ indexes a sample in a ``multiviewed" batch (in this case multiple fonts/augmentations of characters with the same identity), $P(i)$ is the set of indices of all positives in the multiviewed batch that are distinct from $i$, $A(i)$ is the set of all indices excluding $i$, and $z$ is an embedding of a sample in the batch. Training details are describe in the supplementary materials. 

To compute characters' similarity, we embed their image crops, created with a reference font (Google Noto), and compute cosine similarity with a Facebook Artificial Intelligence Similarly Search backend \cite{johnson2019billion}.

Figure \ref{fig:homoglyph} shows representative examples of characters and their five nearest neighbors. Characters with similar vector representations have qualitatively similar appearances. 

\texttt{HOMOGLYPH} shares common elements with EfficientOCR \cite{carlson2023efficient}, an OCR architecture that learns to recognize characters by contrastively training on character crops rendered with augmented digital fonts. Different augmentations of a character provide positive examples. At inference time, localized characters are OCR'ed by retrieving their nearest neighbor from an index of exemplary character embeddings. The OCR application of contrastive learning on character renders aims to retrieve the same character in an offline index, whereas \texttt{HOMOGLYPH} measures similarity across characters. While \texttt{HOMOGLYPH} shares the architecture of the EfficientOCR character recognizer, it does not use the same model weights or training data as EffOCR (which does not support Chinese or Korean and is also trained on labeled crops from historical documents).

\subsection{String Matching Methods}
\citet{dunn1946record} - in one of the first treatments of record linkage - wrote: ``Each person in the world creates a Book of Life. This Book starts with birth and ends with death. Its pages are made up of the records of the principal events in life. Record linkage is the name given to the process of assembling the pages of this Book into a volume.''

Edit distance metrics are widely used for this task \textit{e.g.} \citet{levenshtein1966binary, jaro1989advances, winkler1990string}.
Another common approach computes the cosine similarity between $n$-gram representations of strings \cite{okazaki2010simple}.

There are a variety of ways that character-level visual similarity could be incorporated into record linkage. We follow the literature modifying Levenshtein distance, e.g. \citet{masala}, by using cosine distance in the \texttt{HOMOGLYPH} space as the substitution cost. Insertion and deletion costs are set to one. It is straightforward to scale the insertion and deletion costs using parameters estimated on a validation set, but we focus on performance without any tuned parameters to maintain a purely off-the-shelf, self-supervised implementation. 

We compare matching with homoglyphic edit distance to a variety of other methods. The first comparison is to classic Levenshtein distance (insertions, deletions, and substitutions are all equally costly), to isolate the effect of varying the substitution cost. We also compare to the popular Simstring package, which uses a variety of similarity metrics (Jaccard, cosine, and Dice similarity), computed with 2-gram substrings \cite{okazaki2010simple}. The third comparison is to FuzzyChinese, a widely used package that uses strokes or characters as the fundamental unit for n-gram substring representations (we use the default 3-grams). These are compared using the TF-IDF vectors. The strokes in each character are drawn from an external database \cite{chaizi} covering a subset of the CJK script.

\section{Evaluation Datasets} \label{eval}

To our knowledge, there are not widely used benchmarks for evaluating record linkage for the CJK script.
Hence, we develop evaluation data. 
First, we link a dataset on the customers and suppliers of major Japanese firms, drawn from a 1956 Japanese firm publication \cite{pr}, to a firm index of around 7,000 firms. The index is from the same publication but written in a different font. Supply chains are fundamental to the transmission of economic shocks \cite{acemoglu2016networks,acemoglu2012network}, agglomeration \cite{ellison2010causes}, and economic development \cite{hirschman1958strategy, myrdal1957economic, rasmussen1956studies, bartelme2015linkages, lane2022manufacturing}. Supply chains are challenging to study historically, as they require accurate record linkage. This makes them a particularly relevant test case for downstream applications.

Firm names are localized with LayoutParser \cite{shen2021layoutparser} and then OCR'ed twice, to shed light on whether errors tend to be homoglyphic in popular vision-only OCR and vision-language sequence-to-sequence OCR.  We employ two widely used, open-source OCR engines: PaddleOCR and EasyOCR. 
EasyOCR uses a convolutional recurrent neural network (CRNN) \cite{shi2016end}, with learned embeddings from a vision model serving as inputs to a learned language model. PaddleOCR abandons language modeling, dividing text images into small patches, using mixing blocks to perceive inter- and intra-character patterns, and recognizing text by linear prediction \cite{du2022svtr}. 
Neither engine localizes individual characters.

In a second exercise, we use the dataset examined in \citet{arora2023linking}, which links the same customer-supplier list to a firm directory containing over 70,000 firms \cite{teikoku}. 
Examining this dataset allows a comparison of string matching methods to the OCR-free vision only methods and multimodal methods from \citet{arora2023linking}.
This dataset was created with EfficientOCR \cite{carlson2023efficient} and cannot be re-created with EasyOCR or PaddleOCR because the directory is written vertically, which these engines do not support. We would expect EfficientOCR's character retrieval framework to make homoglyphic errors. Performance across datasets created by three highly diverse OCR architectures is important to extensibility, since database collections have also been constructed with diverse OCR architectures. 


Because creating ground truth data for record linkage is costly, we use synthetically generated data for a third set of evaluations. We render place and firm names using different digital fonts and image augmentations, conducting separate experiments for Traditional Chinese, Simplified Chinese, Japanese, and Korean. For Simplified Chinese, Japanese, and Korean, we draw placenames from the Geonames database \cite{wick2015geonames}. Because Traditional Chinese placenames in Geonames are rare, we instead draw from a list of Taiwanese firms, as Taiwan - unlike Mainland China - still uses Traditional Chinese \cite{TaiwanEcon}. We randomly select two image crops of each entity name, and OCR them using EasyOCR and PaddleOCR. Anywhere from 40\% (Simplified Chinese) to 88\% (Traditional Chinese) of OCR'ed string pairs differ between the two OCR engines. We limit the evaluation dataset to pairs where the two string representations differ.\footnote{The sample size is 20,162 for Simplified Chinese, 66,943 for Traditional Chinese, 86,470 for Japanese, and 48,809 for Korean.} 

\section{Results}  \label{homomatch}

\input{tables/pr_results}

Homoglyphic edit distance outperforms the other string matching methods in all three evaluation exercises - across different OCR engines and languages - typically by an appreciable margin. This illustrates that homoglyphic errors in OCR are common and can be captured with self-supervised vision transformers.

Our first evaluation exercise - with linked Japanese supply chain data - aims to elucidate whether homoglyphic matching is as helpful for linking datasets created with vision-language OCR as for linking datasets created with vision-only OCR, and whether it can similarly be useful for linking datasets created with different OCR architectures. 
We hence separately consider results linking PaddleOCR'ed customers and suppliers to the EasyOCR'ed firm index, vice versa, as well as linking when both are OCR'ed by either PaddleOCR or EasyOCR. 
Homoglyphic edit distance outperforms other string matching methods and does so by a similar margin (around 4 percentage points higher accuracy) regardless of the OCR architecture used. FuzzyChinese has the weakest performance, as expected, since many Japanese characters are not covered in their stroke dictionary. 

\input{tables/clippings_compare}

The primary objective of our second evaluation is to compare homoglyphic matching to the OCR-free vision-only and end-to-end multimodal frameworks developed in \citet{arora2023linking}, using customer-supplier data linked to an extensive index of 70K Japanese firms. Homoglyphic distance outperforms all other string matching methods, with a matching accuracy of 82\%. The \citet{arora2023linking} self-supervised multimodal record linkage model - which employs language-image contrastive pre-training on firm image crop-OCR text pairs (following \citet{radford2021learning}) - outperforms homoglyphic distance, with 85\% matching accuracy. The supervised multimodal model outperforms by a wider margin (95\% accuracy). These methods avoid the OCR information bottleneck by using crops from the original document images. Moreover, the language model can understand different ways of writing the same firm name (\textit{e.g.}, using different terms for corporation). 

The \citet{arora2023linking} supervised vision-only approach, which contrastively trains different views of the same firm's image crops to have similar representations, also outperforms homoglyphic matching (88\% accuracy). While homoglyphs do not fully eliminate the OCR information bottleneck, they do significantly reduce it relative to widely used string matching methods (\textit{e.g.} 75\% accuracy with the Simstring package), with the advantage of not requiring labeled data or image crops. 

\input{tables/synthetic_results}

\texttt{HOMOGLYPH} complements end-to-end deep neural methods. The \citet{arora2023linking} methods cannot be used when researchers lack access to the original document images. Moreover, researchers often lack the compute or technical requirements to work with image data, whereas the vast majority of quantitative social science researchers are comfortable processing strings. On the language side, there are many contexts where using a language model may contribute little. Person or location names - for instance - don't contain much natural language relative to firm names. String matching methods remain the most widely used because they are simple and cheap to use off-the-shelf, and there are contexts where more sophisticated methods may not be feasible or offer large incremental gains. \texttt{HOMOGLYPH} is an extensible way to improve string matching when linking OCR'ed datasets.

\input{figs/errors.tex}

Finally, Table \ref{matching_results_synthetic} reports results with the synthetically generated record linkage dataset, to elucidate the performance of homoglyphic matching across languages using the CJK script. Homoglyphs outperform other string matching methods. The only case where the performance of another method is similar is Simplified Chinese, where the FuzzyChinese package using stroke level $n$-grams performs similarly. The stroke dictionary that underlies FuzzyChinese was crafted for Simplified Chinese, yet homoglyphs can perform similarly with self-supervised methods. On Traditional Chinese, which proliferates in historical documents, homoglyphic edit distance offers a nine percentage point accuracy advantage over FuzzyChinese, illustrating the extensibility advantages of self-supervised methods. The accuracy rates are rather low, but this must be interpreted in the context of the dataset, which only includes paired records where the OCR differs.

Figure \ref{fig:errors} provides an error analysis for the synthetic record linkage exercise. The ground truth string is shown in the first column, PaddleOCR is used to OCR the query (column 2) and EasyOCR is used to OCR the key and provides the correct match (column 3). The matches selected from the key by different string matching methods are shown in columns (4) through (7). 

Panel A shows cases where homoglyphic edit distance selects an incorrect match. This typically occurs when the OCR'ed text has a similar visual appearance to another firm in the index, showing the limits of homoglyphs to fully close the OCR information bottleneck. Panel B shows cases where homoglyphic edit distance selects a correct match, avoiding the wrong strings chosen by other methods through exploiting character visual similarity.


\section{Extending Homoglyphs} \label{ancient}

While this study focuses on the modern CJK script, \texttt{HOMOGLYPH} can be extended to any character set. As a proof of concept, we explore its extensibility to ancient Chinese characters. Like other early forms of human writing, ancient Chinese scripts are highly pictorial relative to modern characters.

Using an existing database of grouped ancient characters from different archaeological sites and periods that correspond to the same concept \cite{Sinica_Academia}, we contrastively learn a metric space where the representations of different views of ancient characters denoting the same concept are nearby. We train on 25,984 character views, as well as the corresponding modern augmented fonts. The dataset includes characters from the Shang Dynasty (1600 BC-1045 BC), the Western Zhou (1045 BC-771 BC), the Spring and Autumn Warring States Era (770 BBC -221 BC), and the Qin-Han Dynasties (221BC - circa third century).\footnote{We exclude images from the Shuowen Jiezi - a book on ancient characters - limiting to the most reliable character renders, which were drawn from archaeological sites.} To illustrate homoglyphs, we create a reference set for the Shang Dynasty, randomly choosing one character for each concept.

Figure \ref{fig:ancient} shows representative examples of homoglyphs, consisting of a character and its five nearest neighbors. The modern character descendant - taken from the database - as well as a short description of the ancient concept are provided. The description draws upon \citet{li2012} as well.  

\input{figs/ancient}

The homoglyph sets are able to capture related abstract concepts noted in the archaeological literature. The first line shows that the concepts of writing, law, learning, and morning (``recording the sun'') are homoglyphs, and the second line shows that characters for different types of officials are homoglyphs, as are characters denoting ``joining.'' The final line shows that history and government official are homoglyphs - underscoring the central role of the government in constituting history - as are characters denoting conquest, tying up, and city center (denoted by a prisoner to be executed by the government, which occurred in the city center). 

Not all concepts within each set are related, but many of the connections above have been noted in an archaeological literature examining how ancient peoples conceptualized the world (\textit{e.g.} \citet{wang2003}). That these meanings can be captured using \textit{vision} transformers is a fascinating illustration of the relationship between images, written language, and meaning in ancient societies.

\section{Limitations} \label{limits}
Using homoglyphs for string matching inherits the well-known limitations of string matching. 
In some cases, OCR destroys too much information for record linkage to be feasible with the resulting strings. 
Even with clean OCR, sometimes language understanding is necessary to determine the correct match. 
Homoglyphs do not address other types of string substitutions, like those that result from enumerator misspellings, although in principle a similar contrastive approach could also be developed to quantify other types of string substitutions. 

More sophisticated methods have been developed as alternatives to string matching. For example, \citet{ventura2015seeing} use a random forest classifier trained on labeled data to disambiguate authors of U.S. patents, applying clustering to the resulting dissimilarity scores to enforce transitivity. 
\citet{arora2023linking} use multimodal methods that combine the image crops of entities and their OCR and also develop a vision-only OCR free linkage method.
Bayesian methods have also been used, \textit{e.g.} \citet{sadinle2014detecting,sadinle2017bayesian}. They offer the advantage of uncertainty quantification - another well-known limitation of string matching - but do not scale well. 

While these methods can offer advantages, they are not always applicable. Researchers may lack access to the original document images, or may lack the compute or technical resources to process images, limiting the use of OCR-free or multimodal approaches. Language models are unlikely to be useful in linking individual names, a common application. Labeled data may be infeasibly costly to create at a sufficient scale for training supervised models. Finally, researchers in disciplines like social science often lack familiarity with machine learning methods, but most are familiar with off-the-shelf string matching packages. String matching methods are also cheap to scale to massive datasets. Simple string matching algorithms are often preferred by practitioners and can be the most suitable tool given the constraints. 

\section{Conclusion} \label{conclude}

Homoglyphic edit distance significantly improves string matching accuracy on OCR'ed documents, by integrating information about character similarity from purely self-supervised vision transformers. It can be implemented using a simple, off-the-shelf string matching package.\footnote{Package available at \url{https://pypi.org/project/HomoglyphsCJK/}.} Learning homoglyphs through self-supervised vision transformers is hhighly extensible, including to low resource settings and settings with many characters. By improving record linkage in such settings - where handcrafted features used to improve record linkage are not available - research on important questions requiring linked data can become more representative of the diversity of human societies. 

\clearpage

\setcounter{table}{0}
\renewcommand{\thetable}{S-\arabic{table}} 
\setcounter{figure}{0}
\renewcommand{\thefigure}{S-\arabic{figure}} 
\setcounter{section}{0}
\renewcommand{\thesection}{S-\arabic{section}}

\begin{center}
    \section*{Supplementary Materials}

\end{center}


\section{\texttt{HOMOGLYPH} Model Details}

\subsection{Encoder}
For both of our applications, we use a DINO pre-trained \cite{caron2021emerging} vision transformer (ViT) as the encoder. Our implementation of the ViT comes from the Pytorch Image Models library (timm) \cite{rw2019timm}. Specifically, we use the \textit{vit\_base\_patch16\_224.dino} model that corresponds to the official DINO-pretained ViT-base model with a patch size of 16 and with input resolution of $224^{2}$. The pretrained checkpoint does not have a classification head. 

\subsection{Loss function}

We use Supervised Contrastive loss \cite{khosla2020supervised} as our training objective, as implemented in the PyTorch Metric Learning library \cite{musgrave2020pytorch}, where the temperature parameter is set to 0.1.

\subsection{Data Augmentation}

We deploy several image augmentations, using transformations provided in the Torchvision library \cite{torchvision2016}. These include Affine transformation (only slight translation and scaling allowed), Random Color Jitter, 
Random Autocontrast, Random Gaussian Blurring, and Random Grayscale. Additionally, we pad the character to make the image square while preserving the aspect ratio of the character render.
We do not use common augmentations like Random Cropping or Center Cropping, to avoid destroying too much information.

For augmenting the skeleton of the rendered character itself, we use a variety of digital fonts to render the images. We use 27 fonts for Simplified Chinese, 17 fonts for Traditional Chinese (for both string matching and ancient Chinese), 62 fonts for Korean, and 14 fonts for Japanese.

\section{Application-specific details}
\subsection{Record Linkage}
\subsubsection{Data}
For each script, the dataset consists of images of characters from the corresponding script rendered with different fonts and augmented during training. The number of characters for each script seen during training is given in Table \ref{training_inference_sizes}. Each character can be considered a "class" to which its digital renders belong. Characters do not need to be seen during training to be considered at inference time, an advantage if users wish to expand the homoglpyph sets (\textit{e.g.} because an OCR engine uses a different character set). We illustrate this empirically by expanding the character set to characters covered by the three OCR engines we explore that were not included in our character ranges used initially for training. 

\input{tables/char_size}

\subsubsection{Batching}
\medskip
\textbf{Without hard-negative mining}

Let $\mathcal{B}$ denote the batch size.
A batch consists of $m$ views of  $\dfrac{\mathcal{B}}{m}$ classes sampled without replacement. When all the views for a class are utilized, all images are replaced and the sampling process without replacement starts again. “Views” of a character are augmented digital renders using the fonts and transformations described above. 

One training epoch is defined as seeing all characters and their m views exactly once.

\medskip

\noindent\textbf{With hard-negative mining} \\
\indent We find $k$ nearest neighbors of each character (or class) on a checkpoint trained without hard negatives. We do this by rendering all characters with a “reference font - Noto Serif CJK font (Tc/Sc/Jp/Ko)”  depending upon the script and finding $k$ nearest neighbors using the above checkpoint. 

We create batches as before, but this time, randomly intersperse all hard negative sets (of size k) in the batches. 

One training epoch is now defined as seeing all characters and their m views and additionally, all characters and their hard negative sets (composed of $k-1$ neighbors) and their m views exactly once.

Table \ref{training_hps} contains the number of epochs we trained each model for. 

\input{tables/training_hp}

\subsubsection{Model Validation}

We split the characters into an 80-10-10 train-val-test set.
We embed the validation images and find the nearest neighbor among the embeddings of digital renders of the universe of characters in the script, rendered with the reference font described above. The top-1 retrieval accuracy is used as the validation metric for the selection of the best checkpoint. We see a peak validation accuracy of 90\% for Japanese, 98\% for Korean, 91\% for Traditional Chinese, and 91\% for Simplified Chinese.

\subsubsection{Other training details}

CJK glyphs are similar across the scripts. To converge faster, for the rest of the languages, we initialize the weights of the encoder with the checkpoint used for the \texttt{HOMOGLYPH} encoder for Japanese - the script with the largest number of characters.
We use AdamW \cite{loshchilov2019decoupled} as the optimizer and Cosine Annealing with Warm Restarts \cite{cosinewarm} as the learning rate schedule. We use the standard Pytorch implementation for both. The relevant hyperparameters are listed in Table \ref{training_hps}. We stop training the models once the validation accuracy stagnates and the checkpoint with the best validation accuracy is chosen as the encoder for each script.

\subsection{Homoglyph Sets}

We allow for the expansion of the character set beyond what is seen in training because different OCR engines use different character dictionaries (a list of characters supported by the engine). We take the union of characters from the character dictionaries of PaddleOCR, EasyOCR, and EfficientOCR. 
For each script, we render all its characters using the script's reference font and embed them using the script-specific \texttt{HOMOGLYPH} encoder. For each character, we then find 800-nearest neighbours (measured by Cosine Similarity between the embeddings) among the set of all renders in the reference set. We store these as a look-up dictionary that contains, for each character in a script, its 800 neighbors and its Cosine Similarity with all of them. This look-up dictionary is used in our modified Levenshtein distance implementation to modify the substitution cost. The dictionaries are available in our GitHub repository \cite{quant_char_sim_repo}. 

Table \ref{training_inference_sizes} contains the number of characters that were used to prepare these sets for each script.

\subsection{Implementing the Modified Levenshtein distance}

We use a standard algorithm to calculate Levenshtein distance that uses dynamic programming \cite{wagfish}. The space and time complexity of the algorithm is $\mathcal{O}(mn)$ where $m$ and $n$ are the lengths of the two strings that are being compared. 

We modify this algorithm by switching the standard substitution cost $\lambda$  between two characters  $a$ and $b$ with  $
\lambda*(1 - CosineSimilarity(u(a),u(b))$. Here $u(a)$ and $u(b)$ are the embeddings of the \texttt{HOMOGLYPH} encoder for the script to which $a$ and $b$ belong.  $\lambda$ is a tunable hyperparameter but for simplicity, we fix it as 1 for the results shown in the paper. We also fixed the addition and deletion cost as 1 but in the implementation provided in our package and our GitHub repository, the costs are tunable hyperparameters. 

\subsection{Ancient Chinese Homoglyphs}
\subsubsection{Data}
The source database \cite{Sinica_Academia} from which we collect the ancient Chinese character crops contains 5,024 concepts, comprised of 25,984 character renderings. Each of these concepts is mapped to a modern character. This enables us to insert digital renders of these modern characters using the same fonts as above (for traditional Chinese) to create more variation. A "class" in this case comprises a character cluster - with both ancient crops and modern digital renders forming the positive samples for a class.

We slightly modify the data augmentation scheme for this application to account for the wide variation in writing styles across centuries. We allow for a slight ($-10$ to $+10$ degree) rotation and also add more transformations tailored to this use case - Random Equalize, Random Posterize, Random Solarize, Random Inversion and Random Erase (randomly erase 0-5\% of the image). We apply all augmentations to the digital renders but only apply Random Affine transformation and Random Inversion to the ancient crops. 

\subsubsection{Batching} 
We use the same sampling and batching process as we did for the modern homoglyph models.
The only difference is in how the hard-negative sets are defined. Instead of one nearest neighbor per concept, for each ancient crop within a concept cluster, we find $k$ nearest neighbors. This gives us as many nearest neighbor sets (hard-negative sets) as ancient crops in our dataset. This allows us to account for the fact that the homoglyphs of a character may differ across different historical periods, spanning millennia.

\subsubsection{Model Validation}

We split the character clusters into train and validation sets (90-10). We then transfer modern renders of the characters from the validation set to the train set.
After this, we randomly transfer 50\% of validation images to training.
Only ancient characters remain in the validation set. We then make a reference set by embedding all the modern renders of our character (using the reference font Noto Serif CJK Tc). We use top-1 accuracy as our validation metric which is defined as the proportion of correct retrievals of the corresponding modern render to each ancient image in the validation set. During training, the model reached a peak validation accuracy of 50\% demonstrating the difficult nature of this task. We use this metric for selecting the best checkpoint for our encoder.

\subsubsection{Other training details}

We again use the AdamW optimizer and Cosine Annealing with Warm Restarts as the learning rate schedule. Relevant Hyperparameters are listed in Table \ref{training_hps}.
We stop training when validation accuracy stagnates.

\subsubsection{Creation of Ancient Chinese Homoglyphs}
The creation of homolgyph sets is analogous to the case of modern characters. Instead of using digital renders from a particular font as the "reference set", we look at the five nearest neighbors of ancient characters within a period. We illustrate homoglyphs using The Shang Dynasty period (1600 BC-1045 BC), the most ancient.

\clearpage
\bibliography{cites,custom}
\bibliographystyle{acl_natbib}

\end{document}


\begin{center}
    \section*{Supplementary Materials}

\end{center}


\section{Technical Details}

\subsection{Encoder}
For both of our applications, we use a DINO pre-trained \cite{caron2021emerging} vision transformer (ViT) as the encoder. We used an implementation of the ViT from the Pytorch Image Models library (timm) \cite{rw2019timm}. Specifically, the \textit{vit\_base\_patch16\_224.dino} model that corresponds to the official DINO-pretained ViT-base model with a patch size of 16 and with input resolution of $224^{2}$ is used. The pretrained checkpoint does not have a classification head. The variety in the character renders is as demonstrated in Figure \ref{fig:font_appendix}.
\subsection{Loss function}

We use the Supervised Contrastive loss \cite{khosla2020supervised} as our training objective as implemented in the PyTorch Metric Learning library \cite{musgrave2020pytorch}, where the temperature parameter is set to 0.1.

\subsection{Data Augmentation}

At the image-level, we used several data augmentations implemented within the Torchvision library \cite{torchvision2016} for training. These included Affine transformation (where only slight translation and scaling was allowed), Random Color Jitter, 
Random Autocontrast, Random Gaussian Blurring, Random Grayscale. Additionally, we padded the character to make the image square while preserving the aspect ratio. 
Note that we do not use common augmentations like Random Cropping to avoid destroying too much information.

For augmenting the skeleton of the rendered character itself, we used a variety of digital fonts to render the images. We used 27 fonts for Simplified Chinese, 17 fonts for Traditional Chinese (for both string matching and ancient Chinese), 62 fonts for Korean, and 14 fonts for Japanese. The font files are available in our repo \footnote{\url{https://github.com/dell-research-harvard/Quantifying-Character-Similarity.git}.}.

\section{Application-specific details}
\subsection{Record Linkage}
\subsubsection{Data}
As described in the main text, for each script, the dataset consisted of characters from the corresponding script rendered with different fonts. The number of characters for each script seen during training is given in Table \ref{training_inference_sizes}. Each character can be considered a "class" and its digital renders are examples of each class. 

\input{tables/char_size}

\subsubsection{Batching and Epoch}
\medskip
\textbf{Without hard-negative mining}

Let $\mathcal{B}$ denote the batch size.
A batch consists of $m$ views of  $\dfrac{\mathcal{B}}{m}$ classes sampled without replacement. When all views are utilized, all the images are replaced and the sampling process without replacement starts again. “Views” of a character are augmented digital renders using the fonts and transformations described above. 

One training epoch is defined as seeing all characters and their m views exactly once.

\medskip

\noindent\textbf{With hard-negative mining} \\
\indent We find $k$ nearest neighbors of each character (or class) on a checkpoint trained without hard negatives. We do this by rendering all characters with a “reference font - Noto Serif CJK font (Tc/Sc/Jp/Ko)”  depending upon the script and finding $k$ nearest neighbors using the above checkpoint. 

We create batches as before, but this time, randomly intersperse all hard negative sets (of size k) in the batches. 

One training epoch is now defined as seeing all characters and their m views and additionally, all characters and their hard negative sets (composed of $k-1$ neighbors) and their m views once.

Table \ref{training_hps} contains the number of epochs we trained each model for. 

\input{tables/training_hp}

\subsubsection{Model Validation}

We split the characters into 80-10-10 percent train-test-val.
We embedded the validation images and find the nearest neighbor among the embeddings of digital renders of the universe of characters in the script rendered with the reference font described above. The top-1 retrieval accuracy was used as the validation metric for the selection of the best checkpoint.

\subsubsection{Other training details}

Since CJK glyphs are very similar across the scripts, in order to converge faster, for the rest of the languages, we initialized the weights of the encoder with the checkpoint used for Japanese.
We used AdamW \cite{loshchilov2019decoupled} as the optimizer and  Cosine Annealing with Warm Restarts \cite{cosinewarm} as the learning rate scheduler. We used the standard Pytorch implementation for both of them. Hyperparameters that were changed from the default settings are listed in Table \ref{training_hps}. We stopped training the models once the validation accuracy stagnated and the checkpoint with the best validation accuracy was chosen as the encoder for each script.

\subsection{Homoglyphs Sets}

For each script, we rendered the characters using the reference font defined above to calculate the 800 nearest neighbors among the universe of renders. 
When we extended our approach to matching across OCR engines, we found that different OCR engines use different character dictionaries. For the final set of characters that we used to find the K-nearest neighbors, we used the union of the OCR dictionaries across engines. Expanding the character sets to accommodate the dictionaries of more OCR engines involved rendering the "unseen" characters at inference time.
Table \ref{training_inference_sizes} contains the number of characters that were finally used to prepare these sets.

\subsection{Implementing the Modified Levenshtein distance}

We use a standard algorithm of Levenshtein distance that uses dynamic programming \cite{wagfish}. The space and time complexity of the algorithm is $\mathcal{O}(mn)$ where $m$ and $n$ are the lengths of the two strings that are being compared. 

We modified the algorithm by switching the standard substitution cost $\lambda$  between two characters  $a$ and $b$ with  $
\lambda*(1 - CosineSimilarity(u(a),u(b))$. Here $u(a)$ and $u(b)$ are the embeddings of the \texttt{HOMOGLYPH} encoder for the script to which $a$ and $b$ belong.  $\lambda$ is a tunable hyperparameter. For simplicity, we fixed it at 1. We also fixed the addition and deletion cost to 1 but in our implementation, all of them are tunable hyperparameters. 

\subsubsection{Additional Matching Accuracy Results}

In Tables \ref{matching_results_synthetic_paddle} and \ref{matching_results_synthetic_easy} we show additional matching accuracy results using a single OCR engine (PaddleOCR and EasyOCR respectively). For PaddleOCR, Homoglyphic Edit Distance still outperforms all other record linkage approaches. For EasyOCR, Simstring (dice and jaccard) perform marginally better than homoglyphic distance while it dominates other methods in Simplified Chinese, Japanese and Korean.  
\input{tables/paddle_gt_table}

\input{tables/easy_gt_table}

\subsection{Ancient Chinese Homoglyphs}
\subsubsection{Data}
The source database \cite{Sinica_Academia} from which we scraped the ancient Chinese character crops contained "clusters" comprising character renditions going back in time. Each of these clusters was mapped to a "modern" character. This enabled us to insert digital renders of these modern characters using the same fonts as above (for traditional Chinese) into these clusters to create more variation. A "class" in this case comprises a character cluster - with both ancient crops and "modern" digital renders forming the "examples".

We slightly modified the data augmentation scheme for this application to account for the wide variation in writing styles across centuries. We allowed for a slight (10-degree) rotation and also added more transformations tailored to this use case - Random Equalize, Random Posterize, Random Solarize, Random Inversion and Random Erase (randomly erase 0-5\% of a the image).

\subsubsection{Batching and Epoch} 
We used the same sampling and batching process for training both with and without offline hard-negative mining.
The only difference is in how the hard-negative sets are defined. Instead of one nearest neighbor per character (or cluster), for each ancient crop within a character cluster, we find $k$ nearest neighbors. We then have as many nearest neighbor sets (hard-negative sets) as there are ancient crops in our dataset. This allows us to account for the fact that the homoglyphs of a character may differ across different periods.

\subsubsection{Model Validation}

We split the character clusters into train and validation sets (90-10). We then transferred modern renders of the characters from the validation set to the train set.
After this, we randomly transferred 50\% of validation images to training.
Only ancient characters thus remained in the validation set. We then made a reference set by embedding all the modern renders of our character (in the reference font Noto Serif CJK Tc). We used top-1 accuracy as our validation metric which is defined as the proportion of correct retrieval of the corresponding modern render to each ancient image in the validation set. We used this metric for selecting the best checkpoint as our encoder.

\subsubsection{Other training details}

We again use the AdamW optimizer and Cosine Annealing with Warm Restarts as the learning rate scheduler. Relevant Hyperparameters are listed in Table \ref{training_hps}.
We stopped training when validation accuracy stagnated.

\subsubsection{Creation of Ancient Chinese Homoglyphs}
The creation of homolgyphs sets, in this case, is analogous to the case of modern characters - instead of using digital renders using a particular font as the "reference set", we look at the nearest neighbors of ancient characters within a period. In the paper, we focus on the most ancient one of them - The Shang Dynasty period (1600 BC-1045 BC). 

\clearpage
\bibliography{cites,custom}
\bibliographystyle{acl_natbib}

%% file: figs/char_font.tex
\begin{figure}[t]
    \centering
    \includegraphics[width=\linewidth]{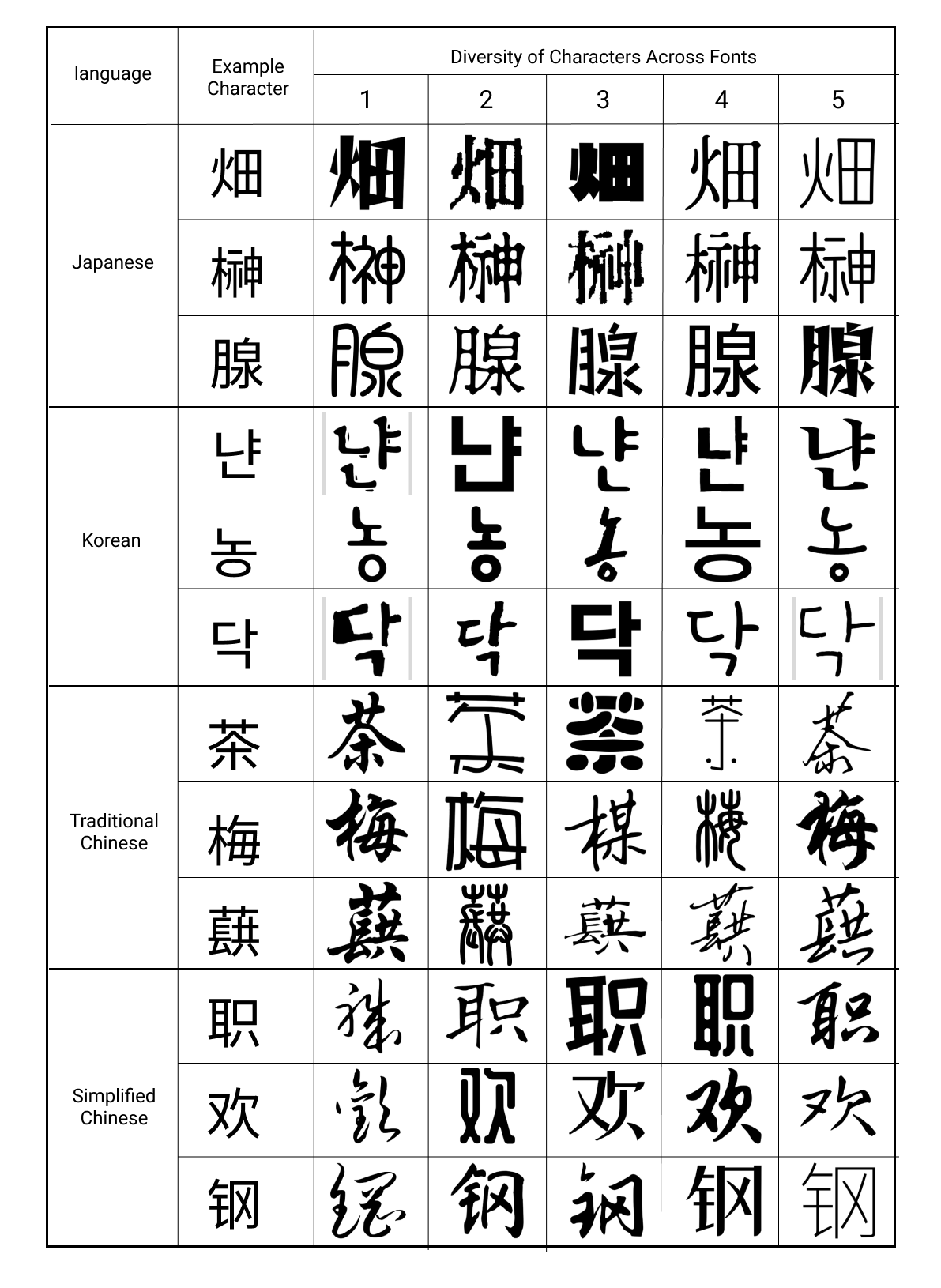}
    \caption{\textbf{Character variation across fonts.} This figure illustrates examples of the same character rendered with different fonts. Augmentations of these comprise positives in the \texttt{HOMOGLYPH} training data.} 
    \label{fig:font}
    \vspace{-4mm}
  \end{figure}

%% file: figs/char_near.tex
\begin{figure}[t]
    \centering
    \includegraphics[width=\linewidth]{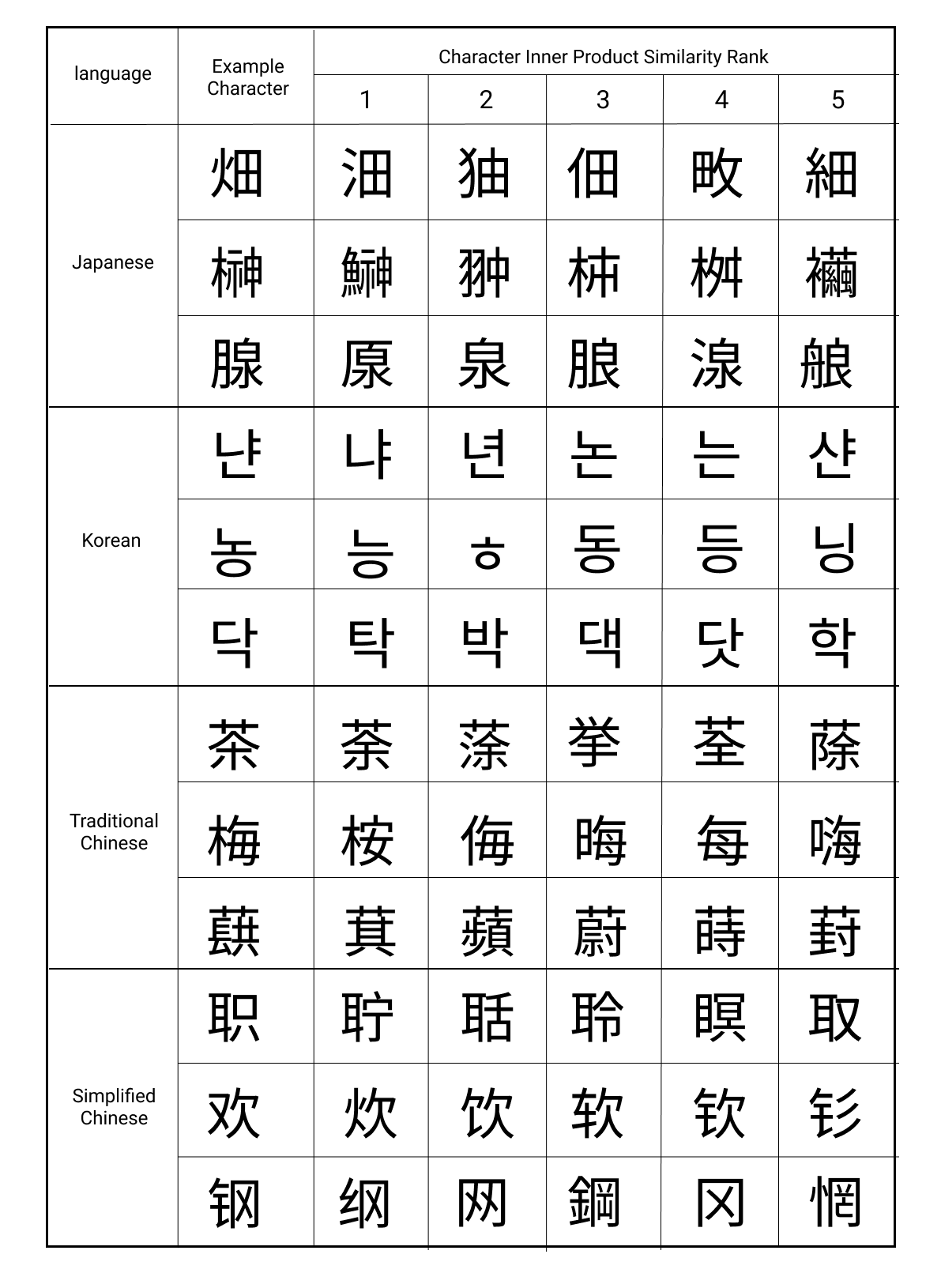}
    \caption{\textbf{Homoglyphs.} This figure illustrates the five nearest neighbors in the \texttt{HOMOGLYPH} embedding space for representative characters.} 
    \label{fig:homoglyph}
    \vspace{-4mm}
  \end{figure}

%% file: tables/pr_results.tex
\begin{table}[t]
    \centering
    \resizebox{\linewidth}{!}{
    \begin{threeparttable}
       \begin{tabular}{lcccc}
      \toprule
      & \multicolumn{4}{c}{OCR Engines} \\
Method	&	Paddle &	Easy &	Paddle &	Easy \\
 & to Easy	& to Paddle & to Paddle & to Easy \\
\cmidrule{1-5}
Homoglyphic 	&	\textbf{0.808}	&	\textbf{0.753}	&	\textbf{0.844}	&	\textbf{0.728}	\\
\ \ distance & \\
Levenshtein 	&	0.766	&	0.697	&	0.807	&	0.693	\\
\ \ distance & \\
Simstring	&	0.762	&	0.662	&	0.787	&	0.673	\\
 \ \  (cosine)\\
Simstring 	&	0.763	&	0.663	&	0.788	&	0.673	\\
\ \ (dice) \\
Simstring 	&	0.763	&	0.663	&	0.788	&	0.673	\\
\ \ (jaccard) \\
FuzzyChinese 	&	0.690	&	0.567	&	0.717	&	0.554	\\
\ \ (stroke) \\
FuzzyChinese 	&	0.533	&	0.445	&	0.559	&	0.464	\\
 \ \ (character) \\
      \bottomrule 
    \end{tabular}
    \end{threeparttable}}
    \caption{\textbf{Baseline Matching Results: Historical Japanese Data}. This table reports accuracy using a variety of different methods for linking Japanese firms from supply chain records to a horizontally written firm directory. The four columns report results when (1) PaddleOCR is used to OCR the firm list and EasyOCR the directory, (2) EasyOCR is used to OCR the firm list and PaddleOCR the directory, (3) PaddleOCR is used to OCR both lists, (4) EasyOCR is used to OCR both lists.}
      \label{matching_results_pr}
\end{table}

%% file: tables/clippings_compare.tex
\begin{table}[t]
    \centering
    \resizebox{\linewidth}{!}{
    \begin{threeparttable}
       \begin{tabular}{lc}
      \toprule
 Method & Accuracy \\
\cmidrule{1-2}

\multicolumn{2}{l}{\textbf{\textit{Panel A: String-Matching}}} \\
Homoglyphic distance & 0.824 \\
Levenshtein distance & 0.731 \\
Simstring (cosine) & 0.748 \\
Simstring (dice) & 0.752 \\
Simstring (jaccard) & 0.752 \\
FuzzyChinese (stroke)  &  0.735 \\ 
FuzzyChinese (character) & 0.618 \\
\multicolumn{2}{l}{\textbf{\textit{Panel B: Neural Methods}}} \\
Self-Supervised Multimodal Linking & 0.849 \\
Supervised Vision-Only Linking & 0.878 \\
Supervised Multimodal Linking  &  \textbf{0.945} \\
      \bottomrule 
    \end{tabular}
    \end{threeparttable}
  }
    \caption{\textbf{Comparisons to fully neural record linkage methods:} This table links Japanese firms from supply chain records to an extensive firm directory. String matching methods are reported in Panel A. End-to-end neural methods are reported in Panel B.}
      \label{clippings_comparison}
\end{table}

%% file: tables/synthetic_results.tex
\begin{table}[t]
    \centering
    \resizebox{\linewidth}{!}{
    \begin{threeparttable}
       \begin{tabular}{lcccc}
      \toprule
	&		&		&	Simplified	&	Traditional	\\
	&	Japanese	&	Korean	&	Chinese	&	Chinese	\\
\cmidrule{1-5}
Homoglyphic 	&	\textbf{0.456}	&	\textbf{0.292}	&	\textbf{0.476}	&	\textbf{0.465}	\\
\ \ distance \\
Levenshtein 	&	0.396	&	0.188	&	0.375	&	0.407	\\
\ \ distance \\
Simstring 	&	0.376	&	0.247	&	0.425	&	0.383	\\
\ \ (cosine) \\
Simstring 	&	0.380	&	0.248	&	0.426	&	0.385	\\
\ \ (dice) \\
Simstring 	&	0.380	&	0.248	&	0.426	&	0.385	\\
\ \ (jaccard) \\
FuzzyChinese	&	0.168	&	0.000	&	0.473	&	0.372	\\
\ \ (stroke) \\
FuzzyChinese 	&	0.230	&	0.110	&	0.137	&	0.197	\\
\ \ (character) \\
    \bottomrule 
    \end{tabular}
    \end{threeparttable}}
    \caption{\textbf{Matching Results: Synthetic Data}. This table reports accuracy linking synthetic paired data generated by OCR'ing location and firm names - rendered with augmented digital fonts - with two different OCR engines.} 
      \label{matching_results_synthetic}
\end{table}

%% file: figs/errors.tex
\begin{figure*}[ht]
    \centering
    \includegraphics[width=.8\linewidth]{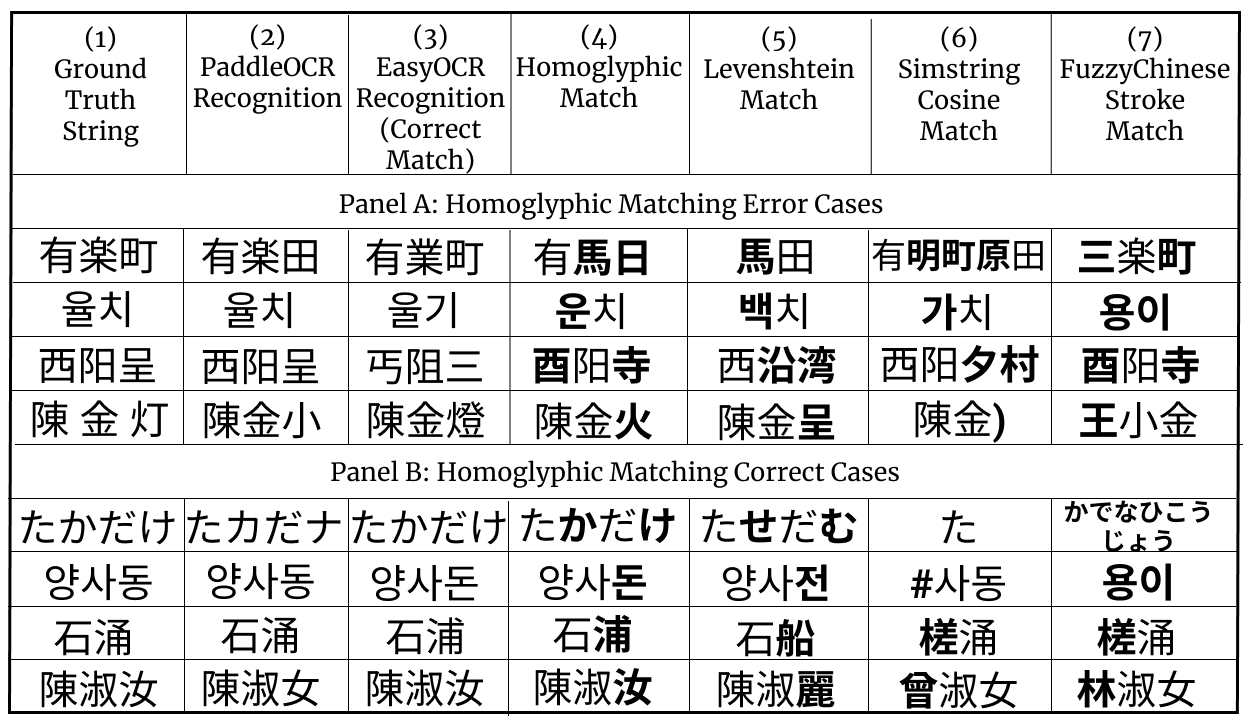}
    \caption{\textbf{Error analysis.} Panel A shows representative errors from homoglyphic matching. Panel B shows representative cases that homoglyphic matching gets correct. The ground truth string is shown in column (1). PaddleOCR is used to OCR the query images (column (2)) and EasyOCR is used to OCR their corresponding keys (column (3)). Columns (4) through (7) give the selected match to the query using different string matching methods, with the correct match shown in column (3). Bold characters differ from the query.} 
    \label{fig:errors}
    \vspace{-4mm}
  \end{figure*}

%% file: figs/ancient.tex
\begin{figure}[t]
    \centering
    \includegraphics[width=\linewidth]{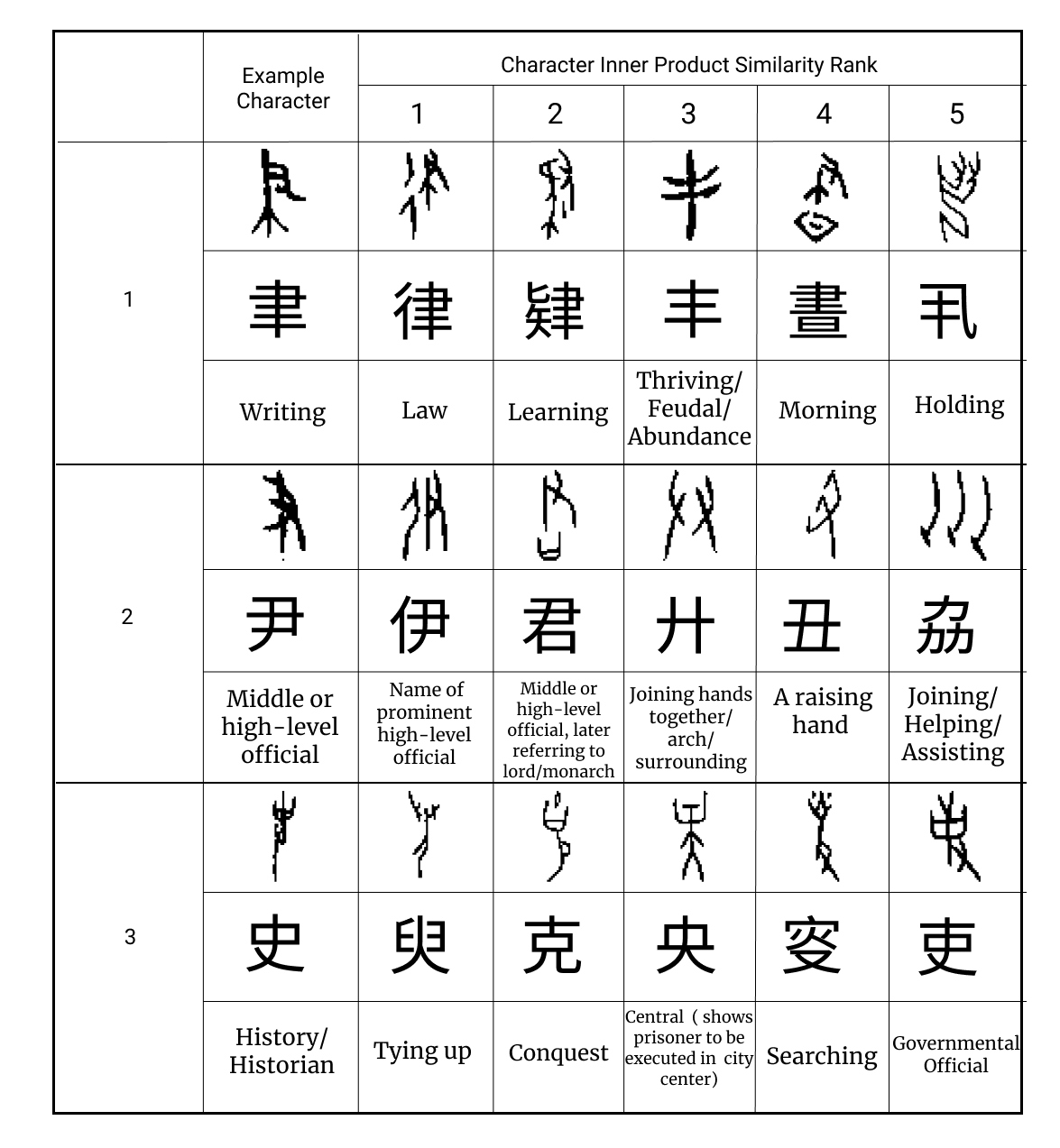}
    \caption{\textbf{Ancient Homoglyphs.} This figure shows homoglyph sets constructed for ancient Chinese, with the descendant modern Chinese character and a description of the character's ancient meaning.} 
    \label{fig:ancient}
    \vspace{-4mm}
  \end{figure}

%% file: tables/char_size.tex
\begin{table}[t]
    \centering
    \begin{tabular}{lcc}
        \toprule
        \textbf{Script} & \textbf{Training} & \textbf{Inference} \\
        \midrule
        Japanese & 17,963 & 17,963 \\
        Simplified Chinese & 6,621 & 7,806 \\
        Traditional Chinese & 8,415 & 8,628 \\
        Korean & 3,686 & 3,729 \\
        \midrule
        \textbf{Total} & \textbf{36,685} & \textbf{38,126} \\
        \bottomrule
    \end{tabular}
    \caption{\textbf{Training and Inference Sizes}. This table shows the training and inference sizes for different language scripts.}
    \label{training_inference_sizes}
\end{table}

%% file: tables/training_hp.tex
\begin{table}[hbt!]
    \centering
    \resizebox{\linewidth}{!}{
    \begin{threeparttable}
       \begin{tabular}{lccccc}
      \toprule
		Model	&	lr	&	
		weight decay	&	T\_0	&	T\_mult	&	Epochs	\\
\cmidrule{1-6}
Japanese - (No HN) 	&	2e-5	&	5e-3	&	1	&	2	&	100	\\
\ \ distance \\
Japanese - (HN) 	&	2e-5	&	5e-3	&	1	&	2	&	30	\\
\ \ distance \\
Simplified Chinese - (No HN) 	&	2e-5	&	5e-3	&	1	&	2	&	30	\\
\ \ distance \\
Simplified Chinese - (HN) 	&	2e-5	&	5e-3	&	1	&	2	&	30	\\
\ \ distance \\
Traditional Chinese - (No HN) 	&	2e-5	&	5e-3	&	1	&	2	&	30	\\
\ \ distance \\
Traditional Chinese - (HN) 	&	2e-5	&	5e-3	&	1	&	2	&	30	\\
\ \ distance \\
Korean - (No HN) 	&	2e-5	&	5e-3	&	1	&	2	&	60	\\
\ \ distance \\
Korean - (HN) 	&	2e-5	&	5e-3	&	1	&	2	&	30	\\
\ \ distance \\
Ancient Chinese - (No HN) 	&	2e-5	&	5e-3	&	300	&	1	&	200	\\
\ \ distance \\
Ancient Chinese - (HN) 	&	2e-5	&	5e-3	&	300	&	3	&	24	\\
\ \ distance \\
    \bottomrule 
    \end{tabular}
    \end{threeparttable}}
    \caption{\textbf{Training Hyperparameters}. This table reports the training hyperparameters used for the models. The lr stands for learning rate, weight decay represents the weight decay factor, T\_0 is the number of steps until the first restart of the learning rate scheduler, T\_mult denotes the factor by which T\_0 is multiplied at each restart, and Epochs indicates the total number of training epochs. Parameters not mentioned here use PyTorch defaults. HN denotes offline hard-negative mining.}
    \label{training_hps}
\end{table}

%% file: tables/paddle_gt_table.tex
\begin{table}[hbt!]
    \centering
    \resizebox{\linewidth}{!}{
    \begin{threeparttable}
       \begin{tabular}{lcccc}
      \toprule
	&		&		&	Simplified	&	Traditional	\\
	&	Japanese	&	Korean	&	Chinese	&	Chinese	\\
\cmidrule{1-5}
Homoglyphic 	&	\textbf{0.649}	&	\textbf{0.520}	&	\textbf{0.577}	&	\textbf{0.584}	\\
\ \ distance \\
Levenshtein 	&	0.577	&	0.356	&	0.492	&	0.515	\\
\ \ distance \\
Simstring 	&	0.552	&	0.333	&	0.574	&	0.513	\\
\ \ (cosine) \\
Simstring 	&	0.556	&	0.334	&	0.574	&	0.514	\\
\ \ (dice) \\
Simstring 	&	0.556	&	0.334	&	0.574	&	0.514	\\
\ \ (jaccard) \\
FuzzyChinese	&	0.232	&	0.002	&	0.569	&	0.497	\\
\ \ (stroke) \\
FuzzyChinese 	&	0.323	&	0.130	&	0.134	&	0.286	\\
\ \ (character) \\
    \bottomrule 
    \end{tabular}
    \end{threeparttable}}
    \caption{\textbf{Matching Results: Synthetic Data(PaddleOCR)}. This table reports accuracy using a variety of different methods for linking location and firm names generated by OCRing digitally augmented fonts with PaddleOCR engines and ground truth strings. Cases where the strings differ are included.The sample size is 8,607 for Simplified Chinese, 61,089 for Traditional Chinese, 67,938 for Japanese, and 15,828 for Korean.}
      \label{matching_results_synthetic_paddle}
\end{table}

%% file: tables/easy_gt_table.tex
\begin{table}[hbt!]
    \centering
    \resizebox{\linewidth}{!}{
    \begin{threeparttable}
       \begin{tabular}{lcccc}
      \toprule
	&		&		&	Simplified	&	Traditional	\\
	&	Japanese	&	Korean	&	Chinese	&	Chinese	\\
\cmidrule{1-5}
Homoglyphic 	&	\textbf{0.632}	&	\textbf{0.547}	&	\textbf{0.612}	&	0.468	\\
\ \ distance \\
Levenshtein 	&	0.578	&	0.397	&	0.522	&	0.425	\\
\ \ distance \\
Simstring 	&	0.552	&	0.353	&	0.509	&	0.473	\\
\ \ (cosine) \\
Simstring 	&	0.556	&	0.356	&	0.510	&	\textbf{0.475}	\\
\ \ (dice) \\
Simstring 	&	0.556	&	0.356	&	0.510	&	\textbf{0.475}	\\
\ \ (jaccard) \\
FuzzyChinese	&	0.175	&	0.000	&	0.545	&	0.463	\\
\ \ (stroke) \\
FuzzyChinese 	&	0.377	&	0.142	&	0.165	&	0.273	\\
\ \ (character) \\
    \bottomrule 
    \end{tabular}
    \end{threeparttable}}
    \caption{\textbf{Matching Results: Synthetic Data(EasyOCR)}. This table reports accuracy using a variety of different methods for linking location and firm names generated by OCRing digitally augmented fonts with EasyOCR engines and ground truth strings. Cases where the strings differ are included.The sample size is 15,911 for Simplified Chinese, 34,254 for Traditional Chinese, 56,292 for Japanese, and 45,416 for Korean.}
      \label{matching_results_synthetic_easy}
\end{table}